\documentclass[conference]{IEEEtran}
\IEEEoverridecommandlockouts
\usepackage{cite}
\usepackage{amsmath,amssymb,amsfonts}
\usepackage{algorithmic}
\usepackage{graphicx}
\usepackage{textcomp}
\usepackage{xcolor}
\def\BibTeX{{\rm B\kern-.05em{\sc i\kern-.025em b}\kern-.08em
    T\kern-.1667em\lower.7ex\hbox{E}\kern-.125emX}}
\usepackage{cite}
\usepackage{amsmath,amssymb,amsfonts}
\usepackage{algorithmic}
\usepackage{graphicx}
\usepackage{textcomp}
\usepackage{xcolor}

\usepackage{graphicx}
\usepackage{tabularx}
\usepackage{amsmath}
\usepackage{multirow}
\usepackage{array}
\usepackage{subcaption}
\usepackage{hyperref}       % hyperlinks
\usepackage{placeins}
\graphicspath{ {./images/} }

\def\BibTeX{{\rm B\kern-.05em{\sc i\kern-.025em b}\kern-.08em
    T\kern-.1667em\lower.7ex\hbox{E}\kern-.125emX}}
\begin{document}

\title{Application of Vision-Language Models to Pedestrian Behavior Prediction and Scene Understanding in Autonomous Driving}

\makeatletter
\newcommand{\linebreakand}{%
  \end{@IEEEauthorhalign}
  \hfill\mbox{}\par
  \mbox{}\hfill\begin{@IEEEauthorhalign}
}
\makeatother

\author{\IEEEauthorblockN{1\textsuperscript{st} Haoxiang Gao\IEEEauthorrefmark{1}\thanks{* Corresponding author.}} 
\IEEEauthorblockA{\textit{ECE Department} \\
\textit{Carnegie Mellon University}\\
Mountain View, United States \\
haoxiang@alumni.cmu.edu}

\and
\IEEEauthorblockN{2\textsuperscript{nd} Li Zhang}
\IEEEauthorblockA{\textit{Computer Science Department} \\
\textit{Columbia University}\\
New York, United States \\
lz2714@columbia.edu}
\and
\IEEEauthorblockN{3\textsuperscript{rd} Yu Zhao}
\IEEEauthorblockA{\textit{Rotman School of Management} \\
\textit{University of Toronto}\\
Toronto, Canada \\
yzqr.zhao@mail.utoronto.ca}
\linebreakand
\IEEEauthorblockN{4\textsuperscript{th} Zhou Yang}
\IEEEauthorblockA{\textit{Department of Statistics} \\
\textit{George Washington University}\\
Washington, DC, United States\\
yangzhou95@hotmail.com}
\and
\IEEEauthorblockN{5\textsuperscript{th} Jinghan Cao}
\IEEEauthorblockA{\textit{Department of Computer Science} \\
\textit{San Francisco State University}\\
San Francisco, United States \\
jcao3@alumni.sfsu.edu}
}

\maketitle

\begin{abstract}
Vision-language models (VLMs) have become a promising approach to enhancing perception and decision-making in autonomous driving. The gap remains in applying VLMs to understand complex scenarios interacting with pedestrians and efficient vehicle deployment. In this paper, we propose a knowledge distillation method that transfers knowledge from large-scale vision-language foundation models to efficient vision networks, and we apply it to pedestrian behavior prediction and scene understanding tasks, achieving promising results in generating more diverse and comprehensive semantic attributes. We also utilize multiple pre-trained models and ensemble techniques to boost the model's performance. We further examined the effectiveness of the model after knowledge distillation; the results show significant metric improvements in open-vocabulary perception and trajectory prediction tasks, which can potentially enhance the end-to-end performance of autonomous driving.
\end{abstract}


\section{Introduction}

Visual reasoning and understanding are crucial in improving autonomous driving by predicting pedestrian behaviors and understanding dynamic environments that ensure the autonomous vehicle performs safe maneuvers. Autonomous driving still faces challenges in handling long-tail safety-critical scenarios in pedestrian interactions, e.g., pedestrians sitting on the roadside, jaywalking, and waiting to cross \cite{taub_2019}\cite{shanks_2023}.
% \textcolor{red}{citations are needed to support this argument}

In autonomous driving, predicting pedestrian behavior typically involves two steps: perception and trajectory prediction. The perception of pedestrians~\cite{weng2023ped3d, li2023pedestrian, zhang2023ped3d} utilizes CNN and Transformer to extract visual and spatial-temporal features from images and point-cloud data and predict pedestrians' bounding boxes and body key points, which has advanced significantly thanks to recent breakthroughs in deep learning and computer vision. Trajectory prediction uses the historical sequences of detected pedestrian bounding boxes and other motion features to forecast their future trajectories. However, most of the work does not leverage visual features and higher-level semantic information from the image to reason about the pedestrian's intents, which requires a superior ability of multi-modal understanding\cite{zhang2023pedestrian}.
% \textcolor{red}{citations. It's better to have a survey paper as the reference}

Foundation models, like Large Language Models (LLMs) and Vision-Language Models (VLMs), which leverage pre-trained knowledge from the web data, show emergent abilities of reasoning and few-shot learning, and have been applied to numerous domains~\cite{huo2024composition, ren2024survey, yang2024optimization, yu2024enhancing}. More and more researchers are starting to use LLMs and VLMs for pedestrian prediction tasks, such as predicting crossing or not crossing~\cite{munirpedvlm, huang2024gpt, gao2024survey}.

Despite recent work, there remains a critical gap in their practical application for autonomous driving. Current approaches lack the capability to:
\begin{itemize}
    \item Domain-Specific Training and Adaptation: Existing LLMs and VLMs are often trained on general-purpose datasets from the Internet, which are quite different from the domain of autonomous driving. Fine-tuning or adapting these models for the specific application of pedestrian behavior understanding remains a challenge.
    \item Efficient model for deployment to vehicles: Deploying large and complex LLMs and VLMs on resource-constrained autonomous vehicles requires more efficient models. Knowledge distillation and model compression techniques are still under development.
    \item Generate Actionable Semantic Signals: LLMs and VLMs often produce descriptions or predictions in natural language but lack the ability to generate structured and actionable signals that can be directly consumed by downstream prediction and planning modules in an autonomous driving system.
\end{itemize}

In this paper, we leverage knowledge distillation\cite{hinton2015distilling} to transfer general knowledge from the pre-trained vision-language foundation model, GPT-4V~\cite{openai2024gpt4technicalreport}, to efficient vision models to extract important semantic attributes, which can predict pedestrian behavior and trajectory. The proposed endeavor manifests the following contributions:

\begin{itemize}
    \item We apply VLM and knowledge distillation to pedestrian behavior prediction and scene understanding tasks, achieving significant improvements in generating diverse and comprehensive pedestrian semantic attributes.
    \item We validated the effectiveness of this approach by open-vocabulary image classification and trajectory prediction tasks and achieved significant metrics improvements compared to baseline models.
    \item We optimize model performance by experiments with multiple vision architectures, pre-trained models, and novel ensemble techniques, which serve as a useful reference and guidance for other researchers and similar applications.
    % \textcolor{red}{The verbs used to claim contributions are too weak}
\end{itemize}

\section{Related Work}

\subsection{Pedestrian Behavior and Intent Prediction}

Pedestrian trajectory and intent prediction have been extensively studied because of the importance of ensuring pedestrian safety. With the advancement of deep learning techniques, LSTM and CNN networks~\cite{alahi2016social, zhao2019multi, salzmann2020trajectron} have been used to capture spatial-temporal patterns in pedestrian motion and make predictions of future trajectories.  Graph neural networks~\cite{mohamed2020social, kosaraju2019social} have also gained more attention, which can model the social relationships in the crowd and interactions between pedestrians and vehicles. Transformer architecture~\cite{ngiam2021scene, nayakanti2023wayformer} has further improved trajectory prediction by using the attention mechanism\cite{vaswani2017attention} to aggregate important information from relevant agents and map elements such as sidewalks, crosswalks, and traffic lights in the scene. Intent prediction often involves estimating high-level behaviors, such as crossing or stopping~\cite{yang2022predicting, zhou2023pit}, by leveraging upstream perception features, such as trajectory history, skeleton key points, and map contexts. Although these methods have made considerable progress, the binary classifications of crossing or not crossing cannot cover the diversity of pedestrian behavior and intents. They also do not take advantage of visual cues in the scene and can not reach a human-like semantic understanding of complex objects and concepts.

\subsection{Vision-Language Models} 
Vision-language models (VLMs) have become a promising approach to improve perception and decision-making in autonomous driving by leveraging both visual and textual data to better understand complex driving scenarios. These models have been used to generate scene descriptions, predict trajectories, and make driving decisions ~\cite{xu2024drivegpt4, chib2024lg, tian2024drivevlm, zhou2024safedrive}. Researchers have begun utilizing VLMs for predicting pedestrian crossings \cite{huang2024gpt}\cite{munirpedvlm}\cite{hamomnipredict}. However, current studies do not fully exploit the capabilities of VLMs to generate comprehensive semantic attributes of pedestrians, which could improve driving decision-making processes. There is limited discussion on how VLMs could enhance downstream prediction and planning tasks and how they could be efficiently deployed to autonomous vehicles with limited computational resources.

\subsection{Knowledge Distillation}
Knowledge distillation effectively optimizes LLMs and VLMs by facilitating the transfer of knowledge from large, computationally expensive teacher models to smaller, more efficient student models without degrading performance ~\cite{liu2023llava, zhu2023minigpt, gu2024minillm}. Similar methods are utilized in the field of autonomous driving, such as improving camera-only perception using the 3D LiDAR teacher model \cite{hong2022cross}, transferring semantic knowledge from 2D to 3D \cite{najibi2023unsupervised}, and optimizing end-to-end driving models \cite{wang2021learning}\cite{feng2024road}.

\section{Methodology}

\subsection{Pedestrian Semantic Labels Extraction}
Existing autonomous driving datasets\cite{nuscene_2024} often fail to provide comprehensive class labels for pedestrian behavior and scene understanding. Many datasets include only basic attributes like ``walking", ``crossing", or ``standing", which lack the detailed semantic information needed for autonomous driving systems that truly understand and interact with pedestrians. To reach human-level perception and decision-making, we need a much deeper and wider understanding of pedestrian intentions, actions, and their relationships with the surrounding environment, e.g. the pedestrian types, body languages, hand signals, and interactions with other agents and map objects.

To address this challenge, we collected detailed annotations using GPT-4V, leveraging its advanced reasoning and natural language generation capabilities, to describe diverse pedestrian behaviors and important aspects of the scene. GPT-4V is prompted with cropped images of the pedestrian and the surrounding environment and the instruction: ``You are a helpful autonomous driving agent. Describe the action and behavior of the pedestrian and the unusual situation in the scene that requires the driver's attention.'' We also provide few-shot examples, to instruct the assistant to answer the question directly, without unnecessary details, and only focus on elements more relevant to autonomous driving. We further apply uni-gram and bi-gram text mining in GPT's annotations after removing common stop words and adverbs to find the most important 256 words and phrases of the highest frequency to describe the pedestrian's behavior, which can be used as semantic labels to train our model to understand the most important pedestrian attributes and scene elements. Examples of annotations are illustrated in Fig. \ref{fig:gpt annotation}

The extracted text labels of pedestrian vocabularies achieve far more coverage than simple pedestrian intents of crossing or not crossing and capture the diversity and complexity of human behavior in traffic scenarios, which can make interactions with pedestrians more intelligent and responsive. There are the following categories for these semantic labels:

\begin{itemize}
    \item Pedestrian types: Autonomous vehicles need to identify different types of pedestrians and understand how they can affect their behaviors. For example, children and the elderly may move unpredictably and slowly, which requires more caution from drivers. Construction workers and figures of authorities might signal hazard areas or direct traffic.

    \item Pedestrian Behaviors: Observing the pedestrian's movement or actions can help predict their future paths, such as moving, waiting, standing, sitting, or walking. Their body postures and hand gestures indicate if they will yield to the vehicles.

    \item The location and surroundings are also crucial. Knowing if a pedestrian is on a crosswalk, waiting at the curb, or walking on the sidewalk helps the autonomous vehicle anticipate its actions.

    \item The weather and environmental conditions significantly influence driving decisions. At night or under poor visibility conditions, extra caution is needed because pedestrians may be less visible or aware of approaching vehicles.

    \item Object interactions: For example, a person with a dog or an elderly with a cane may take longer to cross, or someone using a cell phone may be less attentive to traffic. A person riding a scooter obviously has different movement patterns.

\end{itemize}

\begin{figure}
	\centering
    \includegraphics[width=1.0\linewidth]{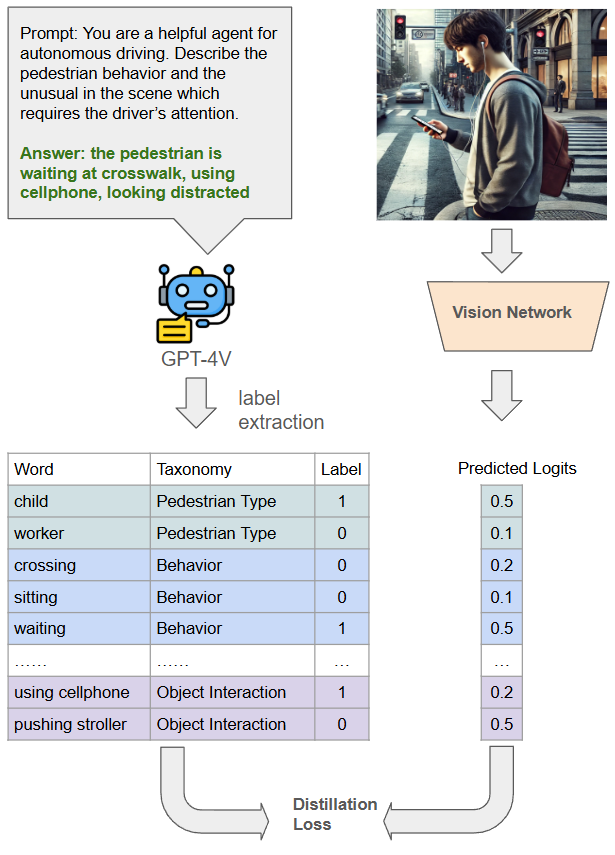}
	\caption{Knowledge distillation architecture leveraging GPT-4V as the teacher model and smaller vision network as the student model.}
	\label{fig:nlp_pipeline}
\end{figure}

\begin{figure}[htb]
     \centering
     \begin{subfigure}[c]{0.22\textwidth}
         \centering
         \includegraphics[width=\textwidth]{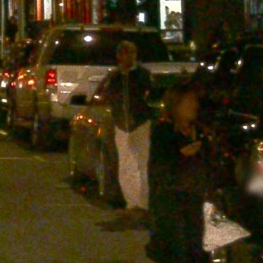}
         % \caption{GPT Annotation: potential jaywalker observing traffic}
         \label{fig: pedestrian observing traffic}
     \end{subfigure}
     \begin{subfigure}[c]{0.22\textwidth}
         \centering
         \includegraphics[width=\textwidth]{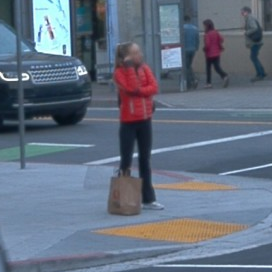}
         % \caption{GPT Annotation: pedestrian waiting at crosswalk}
         \label{fig:waiting}
     \end{subfigure}
    \caption{GPT-4V annotations provide useful semantic information about pedestrian behaviors. GPT annotation for the left image: potential jaywalker observing traffic; GPT annotation for the right image: pedestrian waiting at the crosswalk.}
    \label{fig:gpt annotation}
\end{figure}

\section{Knowledge Distillation to Vision Network}

We formulate the problem as a multi-label classification problem. Given the extracted semantic text labels, we predict the probability that the semantic labels appear in the GPT annotations. For example, the GPT annotation contains ``crossing" and ``walking." The ground truth labels are P(``Crossing") = 1 and P (``Walking") = 1.0, and other labels not in the description are assigned a probability of zero. In this way, we can use semantic labels to supervise our vision network.

\subsection{Knowledge Distillation Method}
\subsubsection{Teacher Model (GPT)}
We will not delve into GPT's internal workings here, as it is a complex black-box transformer model. Assume that it outputs a sequence of text.

We redefine the output as semantic labels that describe the intentions or attributes of the pedestrians of multiple classes $C$. GPT's output texts are converted to class labels with probability:
\begin{align}\label{eq:class_label_prob}
    y_i =
    \begin{cases}
    1, &\text{if i-th class appears in GPT's output}  \\
    0, &\text{otherwise}
    \end{cases}
\end{align}

\subsubsection{Student Model}
We have a lighter-weight vision model to learn the semantic attributes generated by the GPT teacher model. The specific architecture is discussed in the following section. We add a two-layer MLP to adapt the output of the vision network to semantic classes, which takes the form of $f_{\theta}(x)$.

\subsubsection{Binary Cross-entropy loss}
The objective of knowledge distillation is to minimize the difference between the teacher's and the student's predictions. We use a binary cross-entropy loss to predict if the predicted class is present in GPT's answer.
    
\begin{align}\label{eq:bce_loss}
\nonumber
\mathcal{L}(\theta) = -\frac{1}{C} \sum_{i=1}^{C} &[ y_i \log(\sigma(f_{\theta}(x)i)) \\ 
&+ (1 - y_i) \log(1 - \sigma(f{\theta}(x)_i))]
\end{align}

where $\mathcal{L}(\theta)$  is the binary cross-entropy loss, $y_i$ is the teacher's label (0 or 1) for class $i$, $f_{\theta}(x)_i$ is the student's logit for class $i$,  $\sigma$ is the sigmoid function: $\sigma(z) = \frac{1}{1 + e^{-z}}$.

\subsection{Backbone Architecture}
Vision Transformer (ViT)~\cite{dosovitskiy2020vit} and convolutional neural network (CNN)~\cite{lecun1995convolutional} are the most popular backbones for image encoding; we need to consider some trade-offs when choosing between them. ViT excels at capturing global dependencies and contextual relationships within images using the attention mechanism, contributing to richer semantic representations. However, this sacrifices computational efficiency, especially for high-resolution images, because the attention operation has quadratic computational complexity. On the other hand, CNN architecture offers a good balance between performance and efficiency. Its hierarchical structure efficiently processes local features and scales well to larger images. However, it might not capture long-range dependencies as effectively as ViT. We experimented with both backbone architectures for comparisons.

\subsection{Fine-tuning Pre-trained Vision Models}
We leverage vision models pre-trained on large-scale data sets to improve model performance and training speeds.

CLIP~\cite{radford2021learning} is a ground-breaking multi-modal foundation model in natural language processing and computer vision. It uses a contrastive learning approach to associate images with corresponding text captions, bridging the gap between text and image modality. It is widely applied to zero-shot image classification to generate text labels in natural languages. We leverage CLIP's vision embedding, which incorporates the semantic information of images, to accelerate model training and improve accuracy. 
% There are multiple pre-trained image encoder weights of different vision architectures in OpenCLIP~\cite{openai2021clip, openclipgithub}, which is an open-source implementation of CLIP model by OpenAI.
OpenCLIP~\cite{openai2021clip, openclipgithub} is one of the most popular open-source implementations of CLIP model, which provides multiple pre-trained image encoder weights of different vision architectures.

SAM~\cite{kirillov2023segment} and its successor SAM2~\cite{ravi2024sam} are vision foundation models developed by Meta, revolutionizing image and video segmentation. They allow users to select and identify objects within an image or video through different kinds of prompts, such as points, boxes, and text descriptions. The flexible prompt methods and the large training dataset with billions of masks significantly improve the generalization capability to understand the instructions to segment anything. By accurately segmenting pedestrians and other objects in images, SAM can capture essential details regarding their positions and relationships, which can enhance the understanding of pedestrian behaviors. For example, the detailed segmentation can help understand gestures involving different body parts, such as hand and leg movements. It can also improve predicting intents by understanding map objects in the image, such as pedestrians walking at the crosswalk, jaywalking in the middle of traffic, or exiting from the vehicle.

 Sapiens~\cite{khirodkar2025sapiens} is a family of foundation models developed by Meta specifically designed to improve human-centric vision tasks, such as pose estimation, body-part segmentation, and depth estimation, which also makes it a strong candidate to enhance autonomous driving pedestrian prediction. Pre-trained on a massive dataset of over 300 million images, Sapiens already possesses a deep understanding of human appearance, pose, and movement features. By fine-tuning it with autonomous driving scenarios with pedestrians, we can transfer the pre-trained knowledge for better pedestrian behavior and scene understanding.

 We fine-tune these models for 10 epochs with the Adam optimizer and a learning rate of 1e-4 with linear decay.

\subsection{Ensemble of Pre-trained Models}
Different pre-trained vision models have their individual strengths and expertise. For example, CLIP can capture global semantics and interactions, SAM can segment objects of different semantic classes, and Sapiens can leverage domain-specific pre-training for human body data. Yu et al. \cite{liu2024application} and other research in computer vision \cite{lu2025journey} demonstrated that combining deep learning models with an ensemble approach can significantly improve accuracy. These studies have provided valuable guidance for our model design and development. We can further boost the performance by using the ensemble of these models. The overall model architectures of ensembles are demonstrated in Fig. \ref{fig:arch}.

\subsubsection{Mixture-of-Experts Ensemble}
The Mixture of Experts (MoE) \cite{jacobs1991adaptive} is a machine learning technique 
 routing input features to multiple expert models, which handle the input feature space differently. It uses a simple gating function, which plays the role of selecting and combining the outputs of these experts. The gating function outputs a set of weights for all experts, reflecting their relevance to the input. The output of the model ensemble is computed as a weighted sum of the experts' outputs.

The formula of this approach is as follows:
\begin{align}
    g(x) = \text{Softmax}(CNN(x)) \\
E_{ensemble} = \sum_{i=1}^N g_i(x) \cdot E_i(x)
\end{align}

where $g(x)$ is the gating function predicting experts' probabilities, which uses a ResNet-50 backbone with pre-trained ImageNet weights and only fine-tunes the last two layers. $E_i$ is the i-th expert model, and $E_{ensemble}$ is the output of the ensemble.

\subsubsection{Ensemble by Learnable Query}

Despite the simplicity of MoE, we experiment with cross-attention \cite{vaswani2017attention} to extract and combine useful features from multiple domain expert models. We use a learnable latent query and cross-attention to compute weighted outputs. The key and value projection of each expert model output provides better flexibility in learning the ensemble, and the dot product between the learnable query and keys enables more direct interaction with each expert. The definition is as follows:

\begin{align}
E_{ensemble} = \text{Softmax} \left( \frac{Q (W_k E(x) )^T}{\sqrt{d}} \right) W_v E(x)
\end{align}

where $Q$ is a trainable latent query vector of dimension $d$, $E(x) = [E_1(x), ..., E_N(x)]$ is the matrix consisting of embeddings from multiple experts, $W_k$ and  $W_v$ are projection matrices to generate the key and the value of each expert.

\begin{figure}
    \centering
    \includegraphics[width=1.0\linewidth]{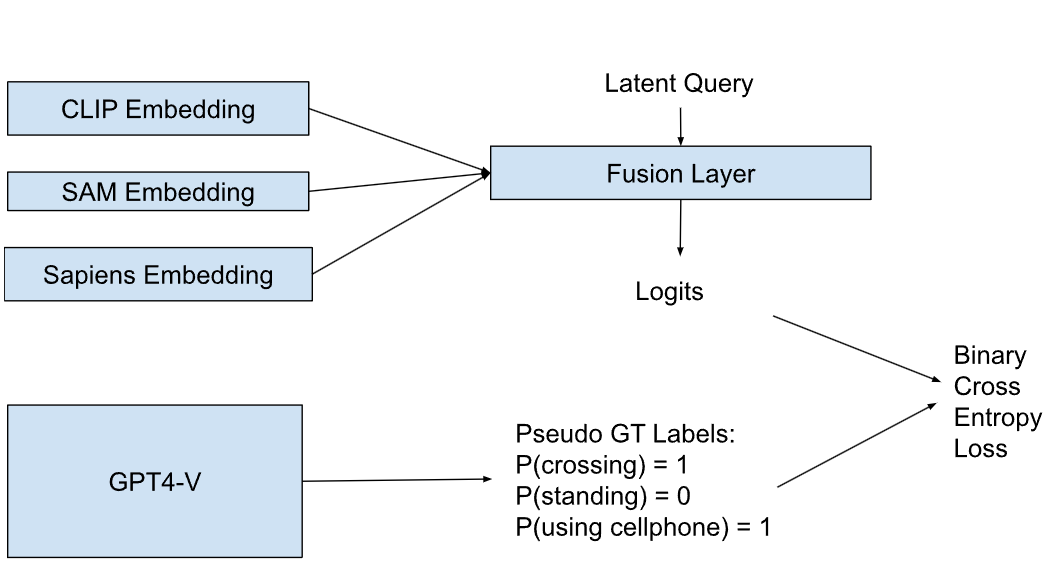}
    \caption{Model architecture of ensemble methods. The left is the ensemble by learnable query, which interacts with each expert model and combines their outputs. The Mixture of Experts is on the right, which determines the routing probabilities given the input feature.}
    \label{fig:arch}
\end{figure}

\subsection{Trajectory Prediction with VLM embeddings}
 The pedestrian behavior signals and semantic text labels can generate early signals for downstream trajectory prediction tasks, and autonomous vehicles can react earlier to avoid collisions. We also propose an experiment to examine the effectiveness of the distilled vision-language embedding in improving the downstream task and potentially enhancing the autonomous vehicle's end-to-end performance. In this task, we use the past 1-second coordinates of pedestrians to predict the position for the next 3 seconds. Similar to other works in trajectory prediction and computer vision \cite{alahi2016social}\cite{salzmann2020trajectron}\cite{dan2024image}\cite{zhao2023collision}, we use a classic RNN trajectory prediction architecture combining extracted image features.

The trajectory prediction task is formulated as below:
\begin{align}\label{eq:trajectory_prediction}
\mathbf{x}_t = [x_t, y_t]
\end{align}

where \(\mathbf{x}_t\) represents the 2D x-y coordinate of the pedestrian at time \(t\).

We use the RNN model with two layers:
\begin{align}\label{eq:hidden_state}
\mathbf{h}_t = f(\mathbf{h}_{t-1}, \mathbf{x}_t \oplus E_{VLM}; \mathbf{W}_h)
\end{align}

 \(\mathbf{h}_t\) is the hidden state at time \(t\), \(f(\cdot)\) is the RNN cell, $E_{VLM}$ is the distilled VLM embedding serving as an additional input feature concatenated with the coordinate feature, and \(\mathbf{W}_h\) represents the learnable parameters of the RNN.

We can predict the future pedestrian trajectory auto-regressively:
\begin{align}\label{eq:rnn_param}
\hat{\mathbf{x}}_{t+1} = g(\mathbf{h}_t; \mathbf{W}_o)
\end{align}

where \(\hat{\mathbf{x}}_{t+1}\) is the predicted next pedestrian coordinate, \(g(\cdot)\) is the linear transformation to output coordinates, \(\mathbf{W}_o\) represents the output layer parameters. The training objective is a Smooth L1 loss to minimize the distance between the predicted coordinates and the ground truth.

\section{Experimental Results}
\subsection{Dataset}

Waymo Open Dataset\cite{sun2020waymodataset} is a widely recognized dataset for researchers to predict and understand pedestrian behavior in the autonomous driving domain. It offers a rich collection of over 1.2 million images, capturing object-centric vehicles and pedestrian instances collected from Waymo's autonomous vehicles in real-world environments. The diversity and completeness of the dataset are crucial for training robust models capable of generalizing across different scenarios. The dataset covers a wide range of pedestrian behaviors, from simple walking to complex interactions with other vehicles. The dataset uses precise 3D bounding box annotations to extract the sequence of images and point cloud data over time. This temporal information is crucial for understanding pedestrian motion patterns and predicting future trajectories.

\subsection{Quantitative Evaluation}

\subsubsection{Open-vocabulary classification comparison with CLIP baseline}

\begin{table*}[ht]
	\centering
    \setlength\tabcolsep{0.15 in}
	\renewcommand{\arraystretch}{1.2}
    \caption{Open-Vocabulary classification evaluation on CLIP model baseline and CLIP model with knowledge distillation (KD) using Precision (P), Recall (R) and F1 Score.}

	\begin{tabular}{|lccc|ccc|ccc|}
		\hline
		\multirow{2}{*}{Methods} & \multicolumn{3}{c|}{Top 1} &\multicolumn{3}{c|}{Top 3}& \multicolumn{3}{c|}{Top 5} \\ \cline{2-10}
		& $P$ & $R$ & $F_1$ & $P$ & $R$ & $F_1$& $P$ & $R$ & $F_1$\\ \hline
		CLIP baseline       & 0.108 & 0.098 & 0.103 & 0.06 & 0.17 & 0.092 & 0.057 &  0.268 & 0.093      \\ \hline
		CLIP w/ KD       & 0.568 & 0.518 & 0.542 & 0.279 & 0.765 & 0.410 & 0.193 & 0.878 & 0.316       \\ \hline

	\end{tabular}
    	\label{tab:zero-shot}
\end{table*}

We compared the CLIP ViT-B model before and after knowledge distillation and evaluated the open-vocabulary classification results as shown in Table \ref{tab:zero-shot}. Because the CLIP model baseline was not fine-tuned on these pedestrian labels, to ensure a fair comparison, we only consider 30 labels of common objects and concepts that appear both in our model's pedestrian-related vocabularies and in ImageNet-21k classes, such as ``vehicle'', ``person'', ``bag'', ``dog'', ``worker'', ``phone'', ``umbrella'', and we calculate metrics based on top-k logits scores outputs from CLIP ViT-B model and the model after knowledge distillation fine-tuning. The evaluation demonstrates significant improvement compared to the CLIP baseline in terms of precision, recall, and F1 score metrics.

Without fine-tuning, CLIP model has difficulty identifying important objects in the scene in the context of autonomous driving, and it was not adapted to the data domain where images are collected in the wild with possible occlusion, blurring, and poor lighting conditions. After knowledge distillation, the model is able to adapt well for the purpose of pedestrian and scene understanding.

\subsubsection{Evaluation on Text Generation}

During inference, the model outputs a set of text labels whose confidence is higher than a pre-set threshold. We manually tune the threshold and use the optimal value of 0.15 to match the average length of GPT-4V's reference answers. We use this threshold to generate text outputs and thereby evaluate the model using common metrics in Natural Language Processing, like BLEU score, precision and recall.

BLEU\cite{papineni2002bleu} score is a popular metric for evaluating the quality of the generated text, and it can be effectively used for assessing the semantic label generation for our use case. It computes the precision of n-grams and includes a penalty for predicted labels that are too short, aiming to balance the precision and recall. To reduce the sensitivity to the word order, we only use uni-gram matching, and the label of phrases, such as ``using cellphone," is treated as one uni-gram.

The evaluation result in Table \ref{table:eval} demonstrates that the CLIP model outperforms SAM and Sapiens, even if Sapiens has the largest number of parameters. The advantage of the CLIP model is its training objective to align text embedding with image embedding and focus on global semantic information of lower-resolution images. While the SAM and Sapiens models excel at traditional computer vision tasks, like segmentation and human pose estimations on higher-resolution images, they lack alignment with the text modality. By leveraging the ensemble of these models, we are able to achieve the best result by selecting the salient information of each model and making the most informed predictions. Our novel method of the learnable-query ensemble also outperforms the existing Mixture-of-Experts method.

We additionally perform experiments utilizing both ViT and CNN architectures, and the comparison shows that CNN, after scaling, can achieve the same performance as ViT with the number of parameters at the same scale.

\begin{table}[htb]
	\centering
    \setlength\tabcolsep{0.15 in}
	\renewcommand{\arraystretch}{1.2}
    	\caption{Evaluation results of knowledge distillation to pre-trained vision models}

	\begin{tabular}{|ccc|}
		\hline
		\centering
		Pre-Trained Model & \# params (M) & BLEU  \\ \hline
	CLIP (ViT-B)  &   88  & 0.274 \\
 	CLIP (ResNet101)  &   44  & 0.262 \\
        CLIP (ResNet50x4) &  100  & 0.285   \\ 
        SAM (ViT-B)       &   89  & 0.254  \\ 
        Sapiens-0.3B      &  336 & 0.214    \\ 
        Ensemble by MoE     &  513 & 0.312   \\ 
        \textbf{Ensemble by learnable query} & \textbf{513} & \textbf{0.326}   \\ 
		\hline
	\end{tabular}
	\label{table:eval}
\end{table}

\subsubsection{Qualitative Evaluation}
We analyzed the examples compared to reference answers generated by GPT. We can observe that the fine-tuned model, after knowledge distillation, was able to describe the actions of the pedestrian in the scene and better understand the context, such as waiting at a bus station, and their more fine-grained actions, such as looking at a cellphone, holding items, and pushing strollers. It can also reason about the weather conditions in the image where the pedestrian holds an umbrella.

We also found in some cases, as shown in Fig. \ref{fig:qualitive_eval_examples}, the model is able to predict more comprehensive semantic attributes of the pedestrian because LLMs tend to use auto-regressive prediction and top-p sampling \cite{openaiapi} to generate the text in answers, which might only focus on the most important aspects of the image but neglect other factors of the scene. By knowledge distillation with a variety of GPT annotations, the fine-tuned model is able to learn the probability distribution of all semantic labels more comprehensively and even provide richer, more complete, and more diverse information about the scene.

\begin{table*}[htb]
\centering
\setlength\tabcolsep{0.3 in}
\renewcommand{\arraystretch}{1.3}
\caption{By adding distilled VLM embedding as an additional input feature, we are able to reduce trajectory error for downstream trajectory prediction task significantly}
\begin{tabular}{|l|c|c|}
\hline
& \begin{tabular}[c]{@{}c@{}}Average Displacement Error \\ (3 seconds, in meters)\end{tabular} & \begin{tabular}[c]{@{}c@{}}Final Displacement Error\\  (3 seconds in meters)\end{tabular} \\ \hline
Baseline   & 0.216 & 0.437 \\ 
Baseline + VLM Embedding & 0.182   & 0.374 \\ \hline
\end{tabular}

\label{tab:pred_eval}
\end{table*}

\subsubsection{Evaluation on Downstream Prediction Task}
We also examine whether VLM embedding can improve performance in the downstream trajectory prediction task, where semantic and image features can provide guidance, such as predicting a shorter trajectory if the pedestrian is waiting and a longer one if the pedestrian is crossing. The evaluation result is shown in Table \ref{tab:pred_eval}, indicating that by leveraging distilled VLM embedding after fine-tuning the CLIP ViT-B model, we are able to significantly reduce trajectory prediction error metrics. Average Displacement Error (ADE) measures the average distance between ground truth and predicted trajectories over all predicted time steps within 3 seconds. Final Displacement Error (FDE) computes the distance between ground truth and predicted trajectories for the final predicted time step.

\begin{figure}[htb]
\centering
\includegraphics[width=0.8\linewidth]{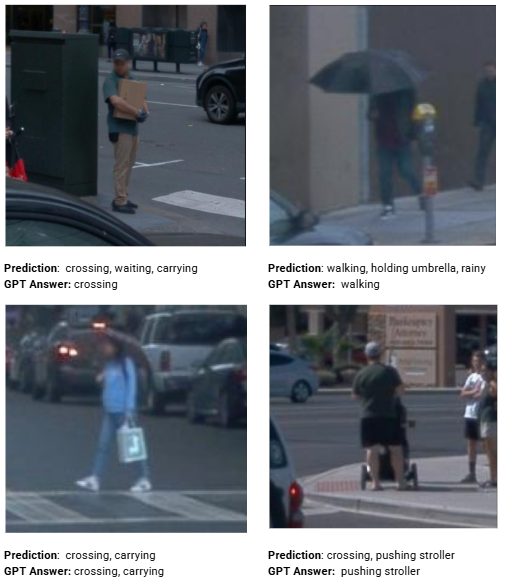}
\caption{Qualitative evaluation shows that our model is able to predict pedestrian behaviors, actions, attentiveness, and infer the weather using objects in the scene. Our model's output is sometimes more comprehensive than GPT-4V's answers.}
\label{fig:qualitive_eval_examples}
\end{figure}
\FloatBarrier

\section{Conclusions}
We propose a knowledge distillation method that distills knowledge from large-scale vision-language foundation models to efficient vision networks, achieving promising results in predicting diverse pedestrian behaviors and semantic attributes. It also improves open-vocabulary perception and trajectory prediction metrics compared to baseline models. We further optimize model performance by conducting experiments with multiple vision architectures, pre-trained models, and ensemble techniques. Our model's predicted behaviors are far beyond the simple pedestrian behavior categories in previous research and cover more semantic attributes essential for pedestrian understanding and interaction. In some cases, our method can correctly generate even more diverse and comprehensive pedestrian behaviors than GPT-4V for autonomous driving.

% \addtolength{\textheight}{-10cm} 

%%%%%%%%%%%%%%%%%%%%%%%%%%%%%%%%%%%%%%%%%%%%%%%%%%%%%%%%%%%%%%%%%%%%%%%%%%%%%%%%

\bibliographystyle{IEEEtran}
\bibliography{IEEEabrv,references}

\begin{thebibliography}{10}
\providecommand{\url}[1]{#1}
\csname url@samestyle\endcsname
\providecommand{\newblock}{\relax}
\providecommand{\bibinfo}[2]{#2}
\providecommand{\BIBentrySTDinterwordspacing}{\spaceskip=0pt\relax}
\providecommand{\BIBentryALTinterwordstretchfactor}{4}
\providecommand{\BIBentryALTinterwordspacing}{\spaceskip=\fontdimen2\font plus
\BIBentryALTinterwordstretchfactor\fontdimen3\font minus \fontdimen4\font\relax}
\providecommand{\BIBforeignlanguage}[2]{{%
\expandafter\ifx\csname l@#1\endcsname\relax
\typeout{** WARNING: IEEEtran.bst: No hyphenation pattern has been}%
\typeout{** loaded for the language `#1'. Using the pattern for}%
\typeout{** the default language instead.}%
\else
\language=\csname l@#1\endcsname
\fi
#2}}
\providecommand{\BIBdecl}{\relax}
\BIBdecl

\bibitem{taub_2019}
\BIBentryALTinterwordspacing
E.~A. Taub, ``How jaywalking could jam up the era of self-driving cars,'' \emph{The New York Times}, Aug 2019. [Online]. Available: \url{https://www.nytimes.com/2019/08/01/business/self-driving-cars-jaywalking.html}
\BIBentrySTDinterwordspacing

\bibitem{shanks_2023}
\BIBentryALTinterwordspacing
A.~Shanks, ``Is it safe to walk in front of an autonomous vehicle?'' Nov 2023. [Online]. Available: \url{https://www.sfexaminer.com/news/transit/waymo-and-cruise-tout-av-safety-around-pedestrians-in-sf/article_687b905a-7a71-11ee-8d29-4fd784af1daa.html}
\BIBentrySTDinterwordspacing

\bibitem{weng2023ped3d}
Z.~Weng, A.~S. Gorban, J.~Ji, M.~Najibi, Y.~Zhou, and D.~Anguelov, ``3d human keypoints estimation from point clouds in the wild without human labels,'' in \emph{Proceedings of the IEEE/CVF Conference on Computer Vision and Pattern Recognition}, 2023, pp. 1158--1167.

\bibitem{li2023pedestrian}
J.~Li, X.~Shi, F.~Chen, J.~Stroud, Z.~Zhang, T.~Lan, J.~Mao, J.~Kang, K.~S. Refaat, W.~Yang \emph{et~al.}, ``Pedestrian crossing action recognition and trajectory prediction with 3d human keypoints,'' in \emph{2023 IEEE International Conference on Robotics and Automation (ICRA)}.\hskip 1em plus 0.5em minus 0.4em\relax IEEE, 2023, pp. 1463--1470.

\bibitem{zhang2023ped3d}
Y.~Zhang, P.~Ji, A.~Wang, J.~Mei, A.~Kortylewski, and A.~Yuille, ``3d-aware neural body fitting for occlusion robust 3d human pose estimation,'' in \emph{Proceedings of the IEEE/CVF International Conference on Computer Vision}, 2023, pp. 9399--9410.

\bibitem{zhang2023pedestrian}
C.~Zhang and C.~Berger, ``Pedestrian behavior prediction using deep learning methods for urban scenarios: A review,'' \emph{IEEE Transactions on Intelligent Transportation Systems}, vol.~24, no.~10, pp. 10\,279--10\,301, 2023.

\bibitem{huo2024composition}
M.~Huo, P.~Ji, H.~Lin, J.~Liu, Y.~Wang, and Y.~Chen, ``Composition vision-language understanding via segment and depth anything model,'' \emph{arXiv preprint arXiv:2406.18591}, 2024.

\bibitem{ren2024survey}
Q.~Ren, Z.~Jiang, J.~Cao, S.~Li, C.~Li, Y.~Liu, S.~Huo, T.~He, and Y.~Chen, ``A survey on fairness of large language models in e-commerce: progress, application, and challenge,'' \emph{arXiv preprint arXiv:2405.13025}, 2024.

\bibitem{yang2024optimization}
H.~Yang, L.~Yun, J.~Cao, Q.~Lu, and Y.~Tu, ``Optimization and scalability of collaborative filtering algorithms in large language models,'' \emph{arXiv preprint arXiv:2412.18715}, 2024.

\bibitem{yu2024enhancing}
H.~Yu, C.~Yu, Z.~Wang, D.~Zou, and H.~Qin, ``Enhancing healthcare through large language models: A study on medical question answering,'' in \emph{2024 IEEE 6th International Conference on Power, Intelligent Computing and Systems (ICPICS)}.\hskip 1em plus 0.5em minus 0.4em\relax IEEE, 2024, pp. 895--900.

\bibitem{munirpedvlm}
\BIBentryALTinterwordspacing
F.~Munir, S.~Azam, T.~Mihaylova, V.~Kyrki, and T.~P. Kucner, ``{PEDVLM}: {PEDESTRIAN} {VISION} {LANGUAGE} {MODEL} {FOR} {INTENTIONS} {PREDICTION},'' 2024. [Online]. Available: \url{https://openreview.net/forum?id=RAX45dcfA2}
\BIBentrySTDinterwordspacing

\bibitem{huang2024gpt}
J.~Huang, P.~Jiang, A.~Gautam, and S.~Saripalli, ``Gpt-4v takes the wheel: Promises and challenges for pedestrian behavior prediction,'' in \emph{Proceedings of the AAAI Symposium Series}, vol.~3, no.~1, 2024, pp. 134--142.

\bibitem{gao2024survey}
H.~Gao, Z.~Wang, Y.~Li, K.~Long, M.~Yang, and Y.~Shen, ``A survey for foundation models in autonomous driving,'' \emph{arXiv preprint arXiv:2402.01105}, 2024.

\bibitem{hinton2015distilling}
G.~Hinton, ``Distilling the knowledge in a neural network,'' \emph{arXiv preprint arXiv:1503.02531}, 2015.

\bibitem{openai2024gpt4technicalreport}
\BIBentryALTinterwordspacing
OpenAI, ``Gpt-4 technical report,'' 2024. [Online]. Available: \url{https://arxiv.org/abs/2303.08774}
\BIBentrySTDinterwordspacing

\bibitem{alahi2016social}
A.~Alahi, K.~Goel, V.~Ramanathan, A.~Robicquet, L.~Fei-Fei, and S.~Savarese, ``Social lstm: Human trajectory prediction in crowded spaces,'' in \emph{Proceedings of the IEEE conference on computer vision and pattern recognition}, 2016, pp. 961--971.

\bibitem{zhao2019multi}
T.~Zhao, Y.~Xu, M.~Monfort, W.~Choi, C.~Baker, Y.~Zhao, Y.~Wang, and Y.~N. Wu, ``Multi-agent tensor fusion for contextual trajectory prediction,'' in \emph{Proceedings of the IEEE/CVF conference on computer vision and pattern recognition}, 2019, pp. 12\,126--12\,134.

\bibitem{salzmann2020trajectron}
T.~Salzmann, B.~Ivanovic, P.~Chakravarty, and M.~Pavone, ``Trajectron++: Dynamically-feasible trajectory forecasting with heterogeneous data,'' in \emph{Computer Vision--ECCV 2020: 16th European Conference, Glasgow, UK, August 23--28, 2020, Proceedings, Part XVIII 16}.\hskip 1em plus 0.5em minus 0.4em\relax Springer, 2020, pp. 683--700.

\bibitem{mohamed2020social}
A.~Mohamed, K.~Qian, M.~Elhoseiny, and C.~Claudel, ``Social-stgcnn: A social spatio-temporal graph convolutional neural network for human trajectory prediction,'' in \emph{Proceedings of the IEEE/CVF conference on computer vision and pattern recognition}, 2020, pp. 14\,424--14\,432.

\bibitem{kosaraju2019social}
V.~Kosaraju, A.~Sadeghian, R.~Mart{\'\i}n-Mart{\'\i}n, I.~Reid, H.~Rezatofighi, and S.~Savarese, ``Social-bigat: Multimodal trajectory forecasting using bicycle-gan and graph attention networks,'' \emph{Advances in neural information processing systems}, vol.~32, 2019.

\bibitem{ngiam2021scene}
J.~Ngiam, B.~Caine, V.~Vasudevan, Z.~Zhang, H.-T.~L. Chiang, J.~Ling, R.~Roelofs, A.~Bewley, C.~Liu, A.~Venugopal \emph{et~al.}, ``Scene transformer: A unified architecture for predicting multiple agent trajectories,'' \emph{arXiv preprint arXiv:2106.08417}, 2021.

\bibitem{nayakanti2023wayformer}
N.~Nayakanti, R.~Al-Rfou, A.~Zhou, K.~Goel, K.~S. Refaat, and B.~Sapp, ``Wayformer: Motion forecasting via simple \& efficient attention networks,'' in \emph{2023 IEEE International Conference on Robotics and Automation (ICRA)}.\hskip 1em plus 0.5em minus 0.4em\relax IEEE, 2023, pp. 2980--2987.

\bibitem{vaswani2017attention}
A.~Vaswani, ``Attention is all you need,'' \emph{Advances in Neural Information Processing Systems}, 2017.

\bibitem{yang2022predicting}
D.~Yang, H.~Zhang, E.~Yurtsever, K.~A. Redmill, and {\"U}.~{\"O}zg{\"u}ner, ``Predicting pedestrian crossing intention with feature fusion and spatio-temporal attention,'' \emph{IEEE Transactions on Intelligent Vehicles}, vol.~7, no.~2, pp. 221--230, 2022.

\bibitem{zhou2023pit}
Y.~Zhou, G.~Tan, R.~Zhong, Y.~Li, and C.~Gou, ``Pit: Progressive interaction transformer for pedestrian crossing intention prediction,'' \emph{IEEE Transactions on Intelligent Transportation Systems}, 2023.

\bibitem{xu2024drivegpt4}
Z.~Xu, Y.~Zhang, E.~Xie, Z.~Zhao, Y.~Guo, K.-Y.~K. Wong, Z.~Li, and H.~Zhao, ``Drivegpt4: Interpretable end-to-end autonomous driving via large language model,'' \emph{IEEE Robotics and Automation Letters}, 2024.

\bibitem{chib2024lg}
P.~S. Chib and P.~Singh, ``Lg-traj: Llm guided pedestrian trajectory prediction,'' \emph{arXiv preprint arXiv:2403.08032}, 2024.

\bibitem{tian2024drivevlm}
X.~Tian, J.~Gu, B.~Li, Y.~Liu, Y.~Wang, Z.~Zhao, K.~Zhan, P.~Jia, X.~Lang, and H.~Zhao, ``Drivevlm: The convergence of autonomous driving and large vision-language models,'' \emph{arXiv preprint arXiv:2402.12289}, 2024.

\bibitem{zhou2024safedrive}
Z.~Zhou, H.~Huang, B.~Li, S.~Zhao, and Y.~Mu, ``Safedrive: Knowledge-and data-driven risk-sensitive decision-making for autonomous vehicles with large language models,'' \emph{arXiv preprint arXiv:2412.13238}, 2024.

\bibitem{hamomnipredict}
\BIBentryALTinterwordspacing
J.-S. Ham, J.~Huang, P.~Jiang, J.~Moon, Y.~Kwon, S.~Saripalli, and C.~Kim, ``Omnipredict: {GPT}-4o enhanced multi-modal pedestrian crossing intention prediction,'' in \emph{Adaptive Foundation Models: Evolving AI for Personalized and Efficient Learning}, 2024. [Online]. Available: \url{https://openreview.net/forum?id=aGI3lSG6PA}
\BIBentrySTDinterwordspacing

\bibitem{liu2023llava}
H.~Liu, C.~Li, Q.~Wu, and Y.~J. Lee, ``Visual instruction tuning,'' in \emph{NeurIPS}, 2023.

\bibitem{zhu2023minigpt}
D.~Zhu, J.~Chen, X.~Shen, X.~Li, and M.~Elhoseiny, ``Minigpt-4: Enhancing vision-language understanding with advanced large language models,'' \emph{arXiv preprint arXiv:2304.10592}, 2023.

\bibitem{gu2024minillm}
Y.~Gu, L.~Dong, F.~Wei, and M.~Huang, ``Minillm: Knowledge distillation of large language models,'' in \emph{The Twelfth International Conference on Learning Representations}, 2024.

\bibitem{hong2022cross}
Y.~Hong, H.~Dai, and Y.~Ding, ``Cross-modality knowledge distillation network for monocular 3d object detection,'' in \emph{European Conference on Computer Vision}.\hskip 1em plus 0.5em minus 0.4em\relax Springer, 2022, pp. 87--104.

\bibitem{najibi2023unsupervised}
M.~Najibi, J.~Ji, Y.~Zhou, C.~R. Qi, X.~Yan, S.~Ettinger, and D.~Anguelov, ``Unsupervised 3d perception with 2d vision-language distillation for autonomous driving,'' in \emph{Proceedings of the IEEE/CVF International Conference on Computer Vision}, 2023, pp. 8602--8612.

\bibitem{wang2021learning}
H.~Wang, P.~Cai, Y.~Sun, L.~Wang, and M.~Liu, ``Learning interpretable end-to-end vision-based motion planning for autonomous driving with optical flow distillation,'' in \emph{2021 IEEE International Conference on Robotics and Automation (ICRA)}.\hskip 1em plus 0.5em minus 0.4em\relax IEEE, 2021, pp. 13\,731--13\,737.

\bibitem{feng2024road}
K.~Feng, C.~Li, D.~Ren, Y.~Yuan, and G.~Wang, ``On the road to portability: Compressing end-to-end motion planner for autonomous driving,'' in \emph{Proceedings of the IEEE/CVF Conference on Computer Vision and Pattern Recognition}, 2024, pp. 15\,099--15\,108.

\bibitem{nuscene_2024}
\BIBentryALTinterwordspacing
NuScene, ``Nuimages,'' 2024. [Online]. Available: \url{https://www.nuscenes.org/nuimages}
\BIBentrySTDinterwordspacing

\bibitem{dosovitskiy2020vit}
A.~Dosovitskiy, ``An image is worth 16x16 words: Transformers for image recognition at scale,'' \emph{arXiv preprint arXiv:2010.11929}, 2020.

\bibitem{lecun1995convolutional}
Y.~LeCun, Y.~Bengio \emph{et~al.}, ``Convolutional networks for images, speech, and time series,'' \emph{The handbook of brain theory and neural networks}, vol. 3361, no.~10, p. 1995, 1995.

\bibitem{radford2021learning}
A.~Radford, J.~W. Kim, C.~Hallacy, A.~Ramesh, G.~Goh, S.~Agarwal, G.~Sastry, A.~Askell, P.~Mishkin, J.~Clark \emph{et~al.}, ``Learning transferable visual models from natural language supervision,'' in \emph{International conference on machine learning}.\hskip 1em plus 0.5em minus 0.4em\relax PMLR, 2021, pp. 8748--8763.

\bibitem{openai2021clip}
OpenAI, ``Clip: Connecting text and images,'' \url{https://openai.com/blog/clip}, 2021, accessed: 2024-05-22.

\bibitem{openclipgithub}
``Github: Openclip,'' \url{https://github.com/mlfoundations/open_clip}, accessed: 2024-05-22.

\bibitem{kirillov2023segment}
A.~Kirillov, E.~Mintun, N.~Ravi, H.~Mao, C.~Rolland, L.~Gustafson, T.~Xiao, S.~Whitehead, A.~C. Berg, W.-Y. Lo \emph{et~al.}, ``Segment anything,'' in \emph{Proceedings of the IEEE/CVF International Conference on Computer Vision}, 2023, pp. 4015--4026.

\bibitem{ravi2024sam}
N.~Ravi, V.~Gabeur, Y.-T. Hu, R.~Hu, C.~Ryali, T.~Ma, H.~Khedr, R.~R{\"a}dle, C.~Rolland, L.~Gustafson \emph{et~al.}, ``Sam 2: Segment anything in images and videos,'' \emph{arXiv preprint arXiv:2408.00714}, 2024.

\bibitem{khirodkar2025sapiens}
R.~Khirodkar, T.~Bagautdinov, J.~Martinez, S.~Zhaoen, A.~James, P.~Selednik, S.~Anderson, and S.~Saito, ``Sapiens: Foundation for human vision models,'' in \emph{European Conference on Computer Vision}.\hskip 1em plus 0.5em minus 0.4em\relax Springer, 2025, pp. 206--228.

\bibitem{liu2024application}
F.~Liu, S.~Guo, Q.~Xing, X.~Sha, Y.~Chen, Y.~Jin, Q.~Zheng, and C.~Yu, ``Application of an ann and lstm-based ensemble model for stock market prediction,'' in \emph{2024 IEEE 7th International Conference on Information Systems and Computer Aided Education (ICISCAE)}.\hskip 1em plus 0.5em minus 0.4em\relax IEEE, 2024, pp. 390--395.

\bibitem{lu2025journey}
B.~Lu, H.-C. Dan, Y.~Zhang, and Z.~Huang, ``Journey into automation: Image-derived pavement texture extraction and evaluation,'' \emph{arXiv preprint arXiv:2501.02414}, 2025.

\bibitem{jacobs1991adaptive}
R.~A. Jacobs, M.~I. Jordan, S.~J. Nowlan, and G.~E. Hinton, ``Adaptive mixtures of local experts,'' \emph{Neural computation}, vol.~3, no.~1, pp. 79--87, 1991.

\bibitem{dan2024image}
H.-C. Dan, Z.~Huang, B.~Lu, and M.~Li, ``Image-driven prediction system: Automatic extraction of aggregate gradation of pavement core samples integrating deep learning and interactive image processing framework,'' \emph{Construction and Building Materials}, vol. 453, p. 139056, 2024.

\bibitem{zhao2023collision}
S.~Zhao, J.~Zhang, C.~He, M.~Huang, Y.~Ji, and W.~Liu, ``Collision-free emergency planning and control methods for cavs considering intentions of surrounding vehicles,'' \emph{ISA transactions}, vol. 136, pp. 535--547, 2023.

\bibitem{sun2020waymodataset}
P.~Sun, H.~Kretzschmar, X.~Dotiwalla, A.~Chouard, V.~Patnaik, P.~Tsui, J.~Guo, Y.~Zhou, Y.~Chai, B.~Caine \emph{et~al.}, ``Scalability in perception for autonomous driving: Waymo open dataset,'' in \emph{Proceedings of the IEEE/CVF conference on computer vision and pattern recognition}, 2020, pp. 2446--2454.

\bibitem{papineni2002bleu}
K.~Papineni, S.~Roukos, T.~Ward, and W.-J. Zhu, ``Bleu: a method for automatic evaluation of machine translation,'' in \emph{Proceedings of the 40th annual meeting of the Association for Computational Linguistics}, 2002, pp. 311--318.

\bibitem{openaiapi}
\BIBentryALTinterwordspacing
OpenAI, ``Openai api.'' [Online]. Available: \url{https://platform.openai.com/docs/api-reference/chat/create}
\BIBentrySTDinterwordspacing

\end{thebibliography}

\end{document}